\documentclass{article}
\usepackage{arxiv}

\usepackage[utf8]{inputenc} 
\usepackage[T1]{fontenc}    
\usepackage{hyperref}       
\usepackage{url}            
\usepackage{booktabs}       
\usepackage{amsfonts}       
\usepackage{nicefrac}       
\usepackage{microtype}      
\usepackage{lipsum}
\usepackage{graphicx}
\graphicspath{ {./images/} }

\usepackage{enumitem}
\usepackage{wrapfig}
\usepackage{multirow}
\usepackage{booktabs}
\usepackage{graphicx}
\usepackage{subcaption}
\usepackage{caption}
\title{SmallWorlds: Assessing Dynamics Understanding of World Models in Isolated Environments}
\author{Xinyi Li, Zaishuo Xia, Weyl Lu, Chenjie Hao, Yubei Chen}

\author{
  Xinyi Li\textsuperscript{1},
  Zaishuo Xia\textsuperscript{1},
  Weyl Lu\textsuperscript{1},
  Chenjie Hao\textsuperscript{1},
  Yubei Chen\textsuperscript{1,2} \\
  \textsuperscript{1}University of California, Davis
  \textsuperscript{2}Open Path AI Foundation \\
}

\begin{document}
\maketitle

\begin{abstract}
Current world models lack a unified and controlled setting for systematic evaluation, making it difficult to assess whether they truly capture the underlying rules that govern environment dynamics. 
In this work, we address this open challenge by introducing the SmallWorld Benchmark, a  testbed designed to assess world model capability under isolated and precisely controlled dynamics without relying on handcrafted reward signals. 
Using this benchmark, we conduct comprehensive experiments in the fully observable state-space on representative architectures including Recurrent State-Space Model, Transformer, Diffusion model, and Neural ODE, examining their behavior across six distinct domains. 
The experimental results reveal how effectively these models capture environment structure and how their predictions deteriorate over extended rollouts, highlighting both the strengths and limitations of current modeling paradigms and offering insights into future improvement directions in representation learning and dynamics modeling.
\end{abstract}

\section{Introduction}

To learn compact representations of the environment's essential information and evolving dynamics, world models~\cite{ha_wm, dyna} are introduced to mimic the cognitive mechanisms of humans that learn to predict, imagine, and reason about the world around them. 
By serving as an internal simulator of environmental evolution, world models enable agents to plan ahead and generalize from limited experience~\cite{dreamerv3}.
The advancement of such models can greatly enhance precision and long-term reasoning~\cite{lecun2022path}, while inaccuracies or poor generalization can severely degrade performance.

Currently, world models are mostly employed within model-based reinforcement learning (MBRL) frameworks~\cite{dyna, lecun_mbrl, dreamerv3}, where they act as efficient and modular surrogates of the real environment, providing predictions of the future environment state to facilitate policy learning and planning.
However, such world models are usually trained on the downstream agent's replay experiences, meaning their performance is highly sensitive to the choice of RL algorithms and the quality of the partially trained policy. 
Also, the prediction of such world models is commonly restricted to short horizons for planning, hardly demonstrating or utilizing their true capacity.
Meanwhile, when the world model is only an ancillary component in the RL pipeline evaluated by the accumulated reward of the policy, its fundamental abilities including prediction stability, interpretability, and physical consistency remain largely unexamined. 

Motivated by these limitations, we introduce the SmallWorld benchmark to evaluate the prediction and dynamics-understanding capabilities of world models.
The SmallWorld benchmark provides a set of canonical physical tasks designed to isolate and examine fundamental physical principles, such as gravity and elastic collision, that can be basic building blocks of complex environments.
Such tasks can be easily converted into fully-observable state spaces whose deviations can be quantitatively evaluated, and can provide diagnostic insights into how these models capture and represent the underlying physical rules. 
Beyond physical reasoning, SmallWorld also includes geometrical understanding tasks that assess world models' ability to maintain spatial structure and geometric coherence over time.

In this paper, we conduct a thorough evaluation of representative world model architectures under long-horizon imagination and physical failure scenarios, providing interpretable comparisons across representative backbone designs including \textit{Recurrent
State-Space Model (RSSM)}~\cite{rssm, dreamerv3}, temporal Transformer~\cite{vaswani2023attentionneed, iris, twm}, \textit{Neural ODE}~\cite{neuroode, mosim}, and diffusion model~\cite{diffusion, df}.
We experiment with tasks in SmallWorld as well as many standard RL tasks such as \textit{Atari}~\cite{atari}, \textit{Go}~\cite{silver2018alphazero}, and \textit{DMControl Suite}~\cite{dmc}, to form a unified testbed encompassing diverse environment dynamics, with a particular focus on physical and logical understanding of world models, providing valuable insights into their fundamental capabilities and potential directions for improvement.

Although achieving stable and accurate long-horizon prediction remains an open challenge, this work aims to push the limit of world models by formalizing the problem of evaluating world models in isolation and introducing a unified testbed. In summary, the SmallWorld benchmark is designed with the following key focuses:

\begin{itemize}[left=0pt]
    \item \textbf{Evaluating world models in isolation.} We separate the world model from conventional RL pipelines, enabling a direct and interpretable assessment of its predictive stability and consistency.

    \item \textbf{The challenge of long-horizon prediction.} We emphasize evaluating world models on their ability to maintain accurate and stable predictions over long time horizons, revealing failure modes that short-horizon RL evaluations often overlook.

    \item \textbf{Diverse environmental dynamics understanding and generalization.} We evaluate world models on tasks encompassing diverse environmental dynamics, allowing systematic analysis of physical and logical understanding.
\end{itemize}


\begin{figure}
    \centering
    \includegraphics[width=0.8\linewidth]{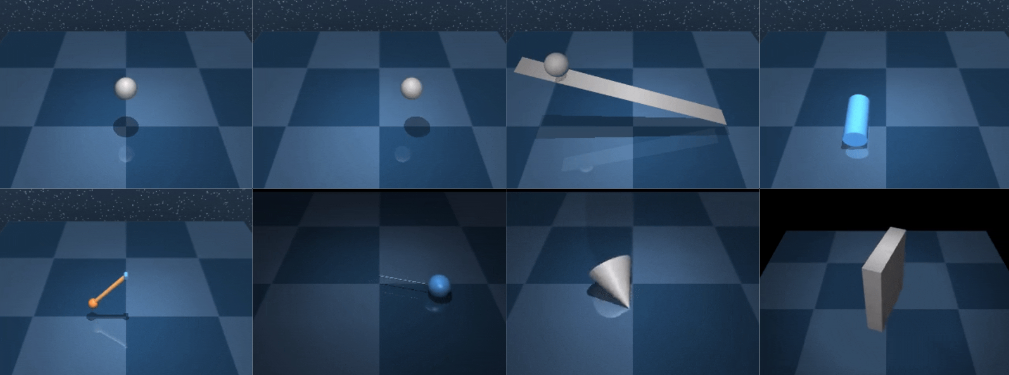}
    \caption{Physical Understanding Tasks Visualization}
    \label{fig:phy}
\end{figure}

\section{Related Work}
\textbf{World Model. } World models typically follow a next-state prediction paradigm: given the current state and an action, predict the subsequent state or observation. 
Recent progress~\cite{bruce2024genie, genie3, mendonca2023structured, he2025matrix, seo2023masked, fang2024towards} has introduced a variety of architectures that enhance latent representation learning~\cite{representation1} and dynamics modeling, such as RNN-based~\cite{rssm, dreamerv1, dreamerv2, dreamerv3} , Transformer-based~\cite{iris, twm, storm}, and Diffusion-based~\cite{df, diamond} dynamic models, achieving strong results on increasingly complex tasks. 
However, the evaluation of these models often relies on accumulated rewards or inconsistent prediction metrics.
In this work, we address this limitation by proposing a unified testbed designed for systematic evaluation and fair comparison.

\textbf{Physical Understanding Evaluation.} Much prior work~\cite{phyinteraction, phyregion, phyre, dynamics3, dynamics4}, has emphasized maintaining physical coherence and reasoning in generative models, yet much of the evaluation remains indirect, often relying on tasks such as solving physics-related puzzles~\cite{phyre} or stacking objects. 
A related work~\cite{howfar} also explores isolated physical dynamics but is conducted in pixel space and limited to three cases: uniform linear motion, elastic collision and parabolic motion.
In this work, we introduce a wider spectrum of tasks that capture broader aspects of physics and perform evaluation in state space for more direct and interpretable analysis.

\textbf{Long-Horizon Prediction.} A line of prior~\cite{mosim, grwm} work has adopted long-horizon prediction as a key indicator of world models' grasp of environment dynamics, typically performing imagination rollouts in highly-coupled environments of certain length and reporting total MSE or visual quality as evidence of improved performance.
In this work, we revisit this setting in isolated environments that expose failure modes more clearly and treat the imagination horizon as a flexible variable that can extend to very long predictions.


\section{The SmallWorld Benchmark}

\subsection{Design Factors}
To push the limit of world model evaluation, we propose the SmallWorld benchmark with the following design factors:

\textbf{Isolated and Controllable Dynamics.}
Existing evaluations of world models are often conducted in environments with complex and highly coupled dynamics, making it challenging to assess whether the model truly understands or follows fundamental environmental rules. For example, Minecraft-based~\cite{minerl} benchmarks are widely adopted for their complexity, but the environment dynamics is composed of diverse factors such as block geometry, gravity, and interactive operations, making it difficult to diagnose which rule is violated when a model's prediction collapses. To address this, we extract common elements that form the basis of complex environments, such as collision and gravity, and evaluate the prediction accuracy of world models under isolated and controlled conditions. This design enables clear and interpretable analysis of world models' areas of strength and weakness.

\textbf{Remove Artificial Reward Function.} 
The use cases of world models often involve reward prediction for downstream planning. However, many open-world tasks do not have certain reward signals, and manually designed reward functions may hinder the interpretability of world models by introducing external bias unrelated to the underlying dynamics. In the SmallWorld benchmark, no artificial reward is assigned or used in evaluation, ensuring that model performance reflects its predictive capability rather than reward modeling.

\textbf{Parameterized Data Distribution} 
An important challenge in training world models is ensuring that they capture the fundamental dynamics of an environment from limited data and generalize to out-of-distribution conditions and unseen scenarios. In this work, we propose tasks with controllable data distributions that can be adjusted through explicit parameters such as the range of initial velocity, while remaining independent of policy selection.

\textbf{Fully Observable States} 
Based on deterministic environment dynamics, we propose tasks that allow access to all information necessary for predicting the next state, for example, the physical states of objects with non-zero degrees of freedom and the forces applied at the current step.
This setup removes confounding factors, such as randomness or missing information, making prediction errors easier to interpret.

\textbf{State-space and Pixel-space inference} 
Much prior work has focused on pixel-space supervision for training world models, using image reconstruction accuracy as a primary evaluation metric. However, pixel space cannot provide complete information required for accurate future-state prediction, such as velocity or out-of-view objects, and it also introduces additional factors such as textures that may interfere with reasoning. The tasks proposed in this work support transformation between state space and pixel space, while most experiments are performed in state space to ensure full observability and enable clearer accuracy measurement and interpretation.

\subsection{Physics Understanding}

Physical understanding is essential for developing advanced world models.
Since physical laws, such as gravity and friction, provide the foundational structure for many complex environments, world models must capture and internalize these laws to achieve accurate simulation in physically grounded settings.
Moreover, the fact that physical rules can be readily formulated mathematically makes them effective tools for analyzing and probing world models.
Failure modes can also be detected when the predictions violate temporal consistency or human's instinct physics understanding.

In this section, we introduce ten tasks for evaluating the capacity of world models for physical understanding. 
As illustrated in Figure~\ref{fig:phy}, the environments are three-dimensional worlds with deterministic and clean physical dynamics, implemented using the physics engine \textit{MuJoCo}~\cite{mujoco}. Each task represents a common physical scenario with interpretable and flexible configurations. 

All interactions occur between rigid bodies with simple geometric shapes, such as spheres and cuboids.
Real number actions ranging in [-1, 1] are scaled forces or torques applied to the primary object. Additionally, no explicit goal or reward function is specified, allowing the environment to mimic open-world scenarios and avoid introducing additional bias. 
The full world state includes all the physical attributes of objects with nonzero degrees of freedom, such as position, orientation, and velocity, making the environment fully observable when paired with the actions.

These tasks can be categorized as follows:

\textbf{Dynamics Governed by Linear Velocity.} \textit{Free Fall Motion}, \textit{Circular Motion}, \textit{Projectile Motion}, \textit{Inclined-Plane Motion}, and \textit{Simple Pendulum Motion} primarily engage translational states driven by linear velocity, making them comparatively easier for models to predict. These dynamics also capture foundational factors such as gravity, that frequently appear in complex environments and are closely tied to instinctive physical reasoning.

\textbf{Dynamics Governed by Rotational Motion.} \textit{Rigid Body Rolling}, \textit{Rigid Body Rotation}, and \textit{Rigid Body Spin} emphasize angular-velocity–dominated dynamics, which are typically harder to anticipate intuitively. These settings introduce more complex motion patterns than purely translational tasks and more fully reflect rigid-body behavior.

\textbf{Dynamics Governed by Energy Conservation.} \textit{Elastic Collision} and \textit{Bouncing Ball Motion} involve dynamics governed by energy conservation, placing direct emphasis on whether the model preserves fundamental conservation principles. These tasks move beyond pure motion prediction and test the model’s ability to maintain invariant quantities.

\subsection{Geometrical Understanding}

\begin{wrapfigure}{r}{0.2\textwidth}
  \vspace{-18pt}
  \begin{center}
    \includegraphics[width=0.2\textwidth]{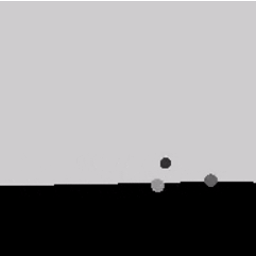}
    \caption{Geometrical Tasks Visualization}
  \end{center}
  \label{fig:geo}
  \vspace{-20pt}
\end{wrapfigure}

In this section, we evaluate geometrical understanding by reformulating the problem into a keypoint reprojection task, where the goal is to predict the pixel-space locations of designated key points in rendered images. This design isolates geometric reasoning by explicitly extracting the transformation from world coordinates to camera coordinates and then to pixel coordinates. 

The ground-truth projection step follows the pinhole camera model adapted to MuJoCo’s coordinate conventions. Each camera has a well-defined extrinsic transform that maps between world and camera frames, and an intrinsic matrix determined by its field of view and resolution. A key point in world space is first converted into the camera coordinate system through the inverse rigid transform and subsequently mapped to pixel space through the intrinsic parameters. 
The world model receives the fully observable state input, including previous pixel coordinates of the key points, their world coordinates, and the current action indicating the camera motion, and then predicts the future coordinates.

Correct projection requires understanding object layout, depth, perspective, camera motion, and visibility constraints, making the task a direct probe of whether the model captures fundamental geometric relationships. Through this formulation, keypoint reprojection serves as a controlled and interpretable measure of a model’s geometric capability and its potential for producing geometrically coherent visual predictions.

\subsection{Combined with Other Benchmarks}
In addition to our proposed task suite, we incorporate several established benchmarks during evaluation to assess a broader spectrum of world model capabilities. These benchmarks can be organized by the type of reasoning or control skill they primarily target: (1) Logical Reasoning: \textit{Go}, (2) Robotic Control: \textit{Panda Hand}~\cite{menagerie}, \textit{DMControl Suite}~\cite{dmc}, (3) Game Control: \textit{Atari Learning Environment (ALE)}~\cite{atari}, (4) Long-term Memory: \textit{Memory-Maze}~\cite{maze}.

One important adaptation is that we do not rely on any reward functions during world model training, and all tasks are converted into fully observable state-space representations, ensuring that learning focuses on predicting environment dynamics rather than exploiting task-dependent rewards.

\section{Experiments}

\subsection{Baseline Architectures}

We experiment with four baseline architectures covering representative dynamic prediction backbones: (1) \textit{Recurrent State-Space Model (RSSM)}, (2) Transformer, (3) Diffusion, (4) Neural ODE. All world models are trained in isolation using the same set of randomly collected episodes, rather than replay buffers generated by a particular agent or policy.

\textbf{Recurrent State-Space Model.} We use the RNN-based RSSM component of \textit{DreamerV3}, a model-based reinforcement learning framework, where deterministic and stochastic latent paths are separately modeled. DreamerV3 has demonstrated strong generalization across different domains and the capability to advance planning in complex environments. In this work, we conduct experiments using the PyTorch implementation of the DreamerV3 world model trained in isolation following the standard configuration.

\textbf{Transformer.} Transformer is a widely used backbone for sequential modeling, and autoregressive architectures allow next-token prediction for imagined rollouts. Following the modeling paradigm of \textit{IRIS}, we employ a autoregressive Transformer as the dynamics model. Each timestep is represented by one observation token and one action token.

\textbf{Diffusion.} We adopt the \textit{Diffusion Forcing} framework, a sequence modeling paradigm that integrates the generative strengths of diffusion models with the sequential nature of autoregressive prediction. By treating sequence generation as a denoising problem with flexible masking, Diffusion Forcing mitigates the error accumulation typically seen in long-horizon rollout. In the context of the SmallWorld benchmark, we adapt this framework for state space prediction. Instead of standard U-Net architectures used in vision, we employ a 2D Transformer as the denoising backbone. 

\textbf{Neural ODE.} We apply \textit{Motion Simulator (MoSim)}, which employs Neural ODE to approximate the explicit dynamic equations of ideal rigid body motion and achieves state-of-the-art prediction accuracy on several physics-grounded tasks. Since MoSim is originally designed for state-space inputs composed of position and velocity components, we follow its default implementation and explicitly specify the dimensions of the position and velocity states for every task.

\begin{figure}[htbp]
\centering
    \begin{subfigure}{0.4\linewidth}  
        \centering
        \includegraphics[width=\linewidth, trim=2cm 1cm 0cm 0cm]{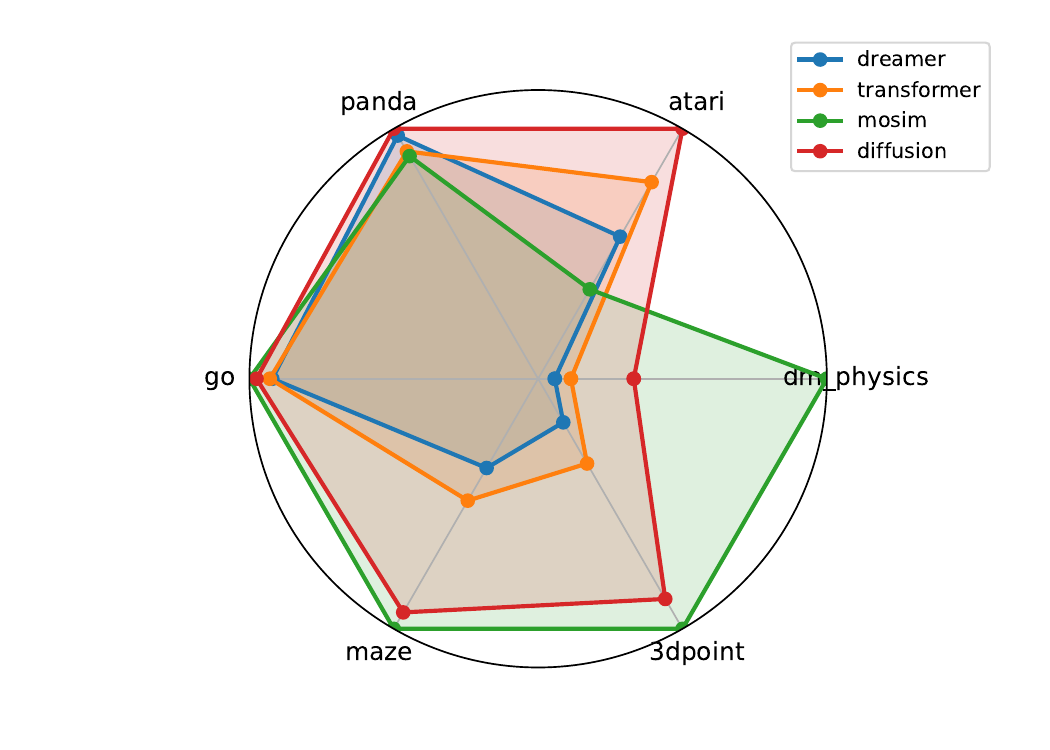} 
        \label{fig:model-a}
    \end{subfigure}
    \begin{subfigure}{0.4\linewidth}
        \centering
        \includegraphics[width=\linewidth, trim=2cm 1cm 0cm 0cm]{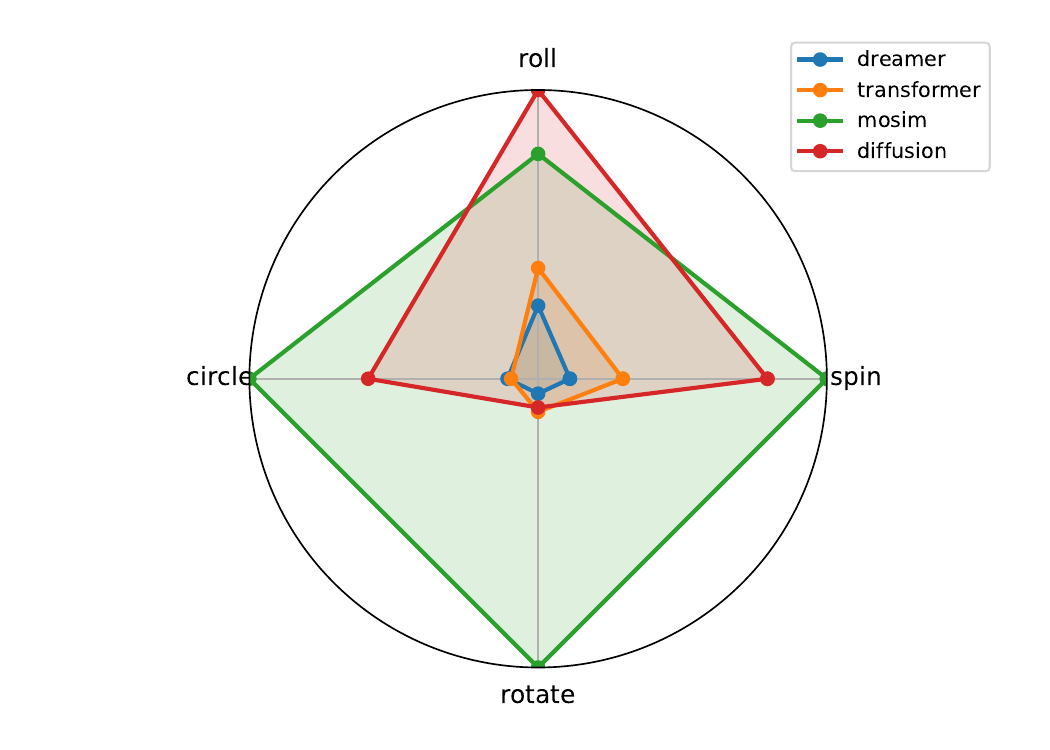}
        \label{fig:model-b}
    \end{subfigure}
\caption{Radar plot illustrating world model capability across task domains, represented as prediction error ratios relative to DreamerV3 and normalized to the maximum value. Left: performance across six task domains; Right: performance on physical reasoning tasks.}
\label{fig:radar}
\end{figure}

\begin{table}[h!]
\centering
\resizebox{\textwidth}{!}{%
\begin{tabular}{lccccccccccc}
\specialrule{1pt}{0pt}{0pt}
\multirow{2}{*}{}
  & \multicolumn{2}{c}{Atari}
  & \multicolumn{4}{c}{SW-Physics}
  & \multicolumn{2}{c}{Panda}
  & \multirow{2}{*}{Go}
  & \multirow{2}{*}{Maze}
  & \multirow{2}{*}{SW-3DPoint} \\
\cmidrule(lr){2-3} \cmidrule(lr){4-7} \cmidrule(lr){8-9}
& Boxing & Pong & Spin & Roll & Circle & Rotate & Push-v1 & Stack-v1 & & & \\
\midrule
RSSM
  & 993.6 & 1094.9 & 341.6 & 0.026 & 0.823 & 0.084 & 0.040 & 0.134 & 0.155 & 2.76 & 4.67 \\
Transformer
  & 820.1 & 545.74 & 64.08 & 0.014 & 0.978 & 0.023 & 0.038 & \textbf{0.085} & 0.153 & 1.75 & 1.669 \\
Mosim
  & 1448.9 & 2683.9 & \textbf{0.632} & 0.0035 & \textbf{0.0012} & \textbf{1.27e-7}
  & 0.030 & 0.095 & \textbf{0.138} & \textbf{0.462} & \textbf{0.091} \\
Diffusion Forcing
  & \textbf{670.0} & \textbf{332.79} & 2.32 & \textbf{0.0018} & 0.018 & 0.0012
  & \textbf{0.024} & 0.227 & 0.143 & 0.537 & 0.148 \\
\bottomrule
\end{tabular}%
}
\caption{Prediction error results, reported as MSE over 90 imagined rollout steps given 10 conditioning steps.}
\label{tab:total}
\end{table}

\subsection{Multi-Domain Capability Evaluation}
Based on the designed benchmark with clean and isolated dynamics, we can assess whether world models capture and internalize the underlying rules and transitions of the environment based on their predictions, more precisely the state-space output based on given action sequence without ground-truth observations.
In this section, we show evaluation results on all architectures introduced in Section 4.1 across the six independent domains described in Section 2, proving a comprehensive and comparable analysis on the overall capacity and areas of weakness of different architectures.

Table~\ref{tab:total} summarizes the quantitative evaluation results across six representative benchmarks, reporting the Mean Squared Error in the state space under best imagination performance of predicting consecutive 90 steps conditioned on 10 ground-truth steps as warm-up phase, which is important for models that rely on historical latent states, such as RSSM.

Across all evaluated tasks, Diffusion Forcing and MoSim consistently achieve the lowest prediction error, indicating superior long-horizon modeling capability compared to the other two architectures. 
Meanwhile, the Transformer architecture performs consistently better than RSSM, suggesting that explicit temporal modeling and attention-based representations provide advantages over complicated latent-state inference when scaling to diverse and complex dynamics.
These results imply that architectures with stronger sequence modeling capacity and implicit noise handling tend to maintain more stable state predictions over extended time horizons. 

The performance also varies considerably across domains as prediction accuracy improves. MoSim achieves high precision on physically grounded tasks in the SmallWorld benchmark but performs poorly on Atari. 
A plausible explanation is that Neuro-ODE models the change in states rather than the states themselves, making it well suited for domains where state transitions follow formalized physical rules, but less effective when state evolution tends to be abrupt and irregular.

Diffusion demonstrates more consistent performance across domains compared with MoSim, achieving the best prediction performance on Atari and several physical reasoning benchmarks. This advantage suggests that its denoising-based iterative refinement provides stronger robustness under complex, irregular state transitions. However, Diffusion shows notable lower precision than MoSim on tasks where extremely fine-grained accuracy is required.
A plausible explanation is that the stochastic noise injection and denoising process introduce small but accumulated deviations, limiting its ability to match the fine-grained accuracy achievable by continuous-time dynamics modeling. This observation highlights a trade-off between robustness and precision inherent to diffusion-based world models.

\subsection{Long-horizon Imagination}

\begin{figure}
  \centering
  \includegraphics[width=\linewidth]{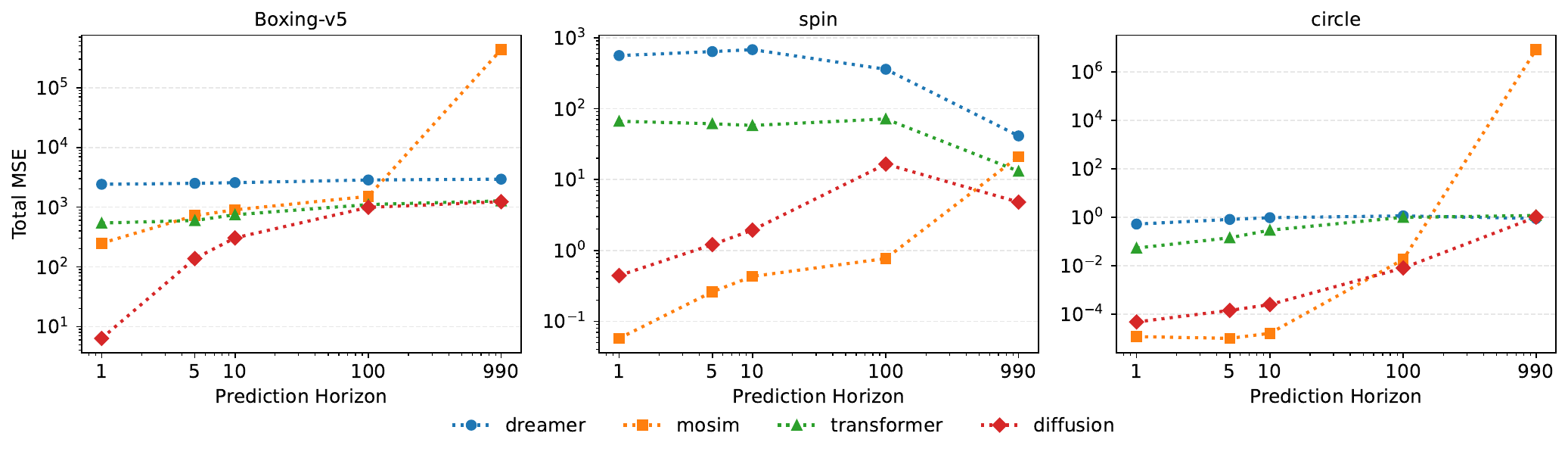}
  \caption{Prediction error as a function of imagination horizon across three representative tasks.}
  \label{fig:tvar}
\end{figure}

Maintaining stable and accurate predictions as the prediction horizon increases is a strong indicator of whether a world model has internalized the underlying environment dynamics. 
In this section, we investigate how several world models behave under increasing rollout lengths, and Figure~\ref{fig:tvar} illustrates their error growth across three representative tasks.

One notable pattern we observe is that although MoSim achieves relatively low short-term error, its predictions become unstable and exhibit exponential error growth as the prediction horizon increases. This suggests substantial error accumulation and weak self-correction, since MoSim generates the next state solely from the immediately preceding one.
Meanwhile, the other three models demonstrate smaller difference in total error as horizon expands, one plausible explanation is that they gradually lose information along the imagination trajectory and eventually rely only on their fundamental understanding of the underlying dynamics.
We also observe a decrease in error on the spin task, likely because the object eventually settles into a static state after its initial rotational motion, making subsequent predictions substantially easier.


\section{Discussion and Conclusion}
In this work, we addressed the open problem of how to rigorously evaluate world models and introduced SmallWorld, a unified benchmark designed to assess world models in isolation, without the confounding effects of reward design or policy-driven data. By constructing tasks with deterministic, controllable, and fully observable dynamics, we provided a principled setting for examining whether a model has internalized the underlying rules such as physical and geometric evolution. Through extensive experiments across six domains and four representative architectures, we demonstrated that this form of isolated evaluation reveals structural differences that conventional RL-centric evaluations often obscure. The results show that MoSim achieves the highest precision on structured physical dynamics, while Diffusion Forcing displays superior robustness in irregular and visually complex domains such as Atari, highlighting the discriminative power of the benchmark in exposing architectural inductive biases.

We hope this framework serves as a step toward a more principled methodology for world model evaluation, offering a controlled testbed that encourages progress in dynamics understanding, representation learning, and long-horizon prediction. Future work may extend this paradigm to richer physical systems, partially observable settings, and visually grounded environments, as well as broader families of architectures, ultimately advancing the development of world models that reason, extrapolate, and generalize with greater reliability.


\newpage
\bibliographystyle{unsrt}
\bibliography{main}

\newpage
\appendix
\section{Physical Understanding Task Setup}
In this section, we introduce the physical understanding tasks in detail, providing their design motivation and setup in the Mujoco simulator.

\textbf{Elastic Collision.} This task involves elastic collisions between a ball and a static wall, or between two moving balls. In an ideal elastic collision, kinetic energy should be conserved and no energy is lost through deformation or friction. Prediction accuracy is assessed based on deviations in kinetic energy before and after the collision. Non-ideal behaviors such as partial energy loss or angular deflection can also be introduced to evaluate robustness and generalization.

\textbf{Bouncing Ball Motion.} This task involves a rigid ball moving in a frictionless horizontal plane. The ball is confined within a square environment bounded by four static vertical walls and is initialized with a non-zero horizontal velocity. During motion, two physical principles should hold: (1) the ball maintains uniform linear motion when no collisions occur, and (2) when colliding with a wall, the reflection angle equals the incidence angle, consistent with an ideal elastic collision model. Prediction accuracy is evaluated based on deviations in ball position, velocity, and reflection behavior relative to these expected rules.

\textbf{Free Fall Motion.} Gravity is one of the most fundamental constraints governing many physical environments. In this task, the objective is to evaluate whether the world model can correctly capture and predict motion governed solely by gravity in an isolated setting, serving as a foundation for more complex dynamics. The scene consists of a rigid ball undergoing free fall under gravitational acceleration, followed by elastic bouncing behavior upon colliding with the floor.

\textbf{Circular Motion.} This task involves a rigid ball attached to a central point by a flexible string. The ball moves on a smooth, frictionless plane. The control action applies a tangential force along the direction of the instantaneous velocity of the ball, modifying its speed while the string tension supplies the required centripetal force. Together, these forces allow the ball to maintain its circular trajectory. 

\textbf{Projectile Motion.} In contrast to the free-fall motion, this task involves a ball with a non-zero initial horizontal velocity, serving as an extension of the free-fall task. After release, the ball follows a parabolic trajectory under the influence of gravity. The motion can be decomposed into uniform horizontal motion and uniformly accelerated vertical motion, forming a classical example of two-dimensional kinematics.

\textbf{Inclined-Plane Motion.} This task examines the motion of a rigid ball that descends along a fixed inclined surface. Subject to gravitational acceleration and an adjustable frictional coefficient, the expected dynamic includes a steady acceleration component aligned with the plane. Accurate predictions should exhibit a monotonic variation in position and velocity along the inclined direction.

\textbf{Simple Pendulum Motion.} This task focuses on the periodic motion of a simple, classic pendulum that oscillates under gravity in a two-dimensional plane. Valid world model predictions should capture consistent oscillation timing, stable amplitude behavior, and smooth transitions at turning points.

\textbf{Rigid Body Rolling.} This task considers the motion of a horizontally oriented cylindrical body that rolls on a flat surface. The cylinder is initialized with a non-zero angular velocity around its central axis. Under ideal rolling conditions, the no-slip constraint dictates a fixed relationship between linear velocity and angular speed, and the system’s kinetic energy is split between rotation and translation in a predictable ratio, providing a focused assessment of whether the world model can capture the dynamics of rotation–translation coupling.

\textbf{Rigid Body Rotation.} This task involves a single rigid object that undergoes ideal uniform rotation about its central axis. The object is initialized with a non-zero angular velocity aligned with its axis of symmetry. With friction and external forces excluded, the rotational motion should remain perfectly steady such that the angular velocity stays constant, and the orientation evolves at a rate determined by this constant speed. This setup evaluates whether the world model can reliably reproduce pure rotational dynamics and maintain stable orientation updates over time.

\textbf{Rigid Body Spin.} This task features an conical object to approximate the behavior of a spinning top. The object is initialized with a non-zero angular velocity around its vertical axis, creating an initially stable spinning state. As friction gradually acts on the system, the top begins to lose rotational stability and its axis of rotation slowly diverges away from the upright configuration. The motion eventually transitions from steady spin to wobbling and finally settles into a static resting state.

\end{document}